\pdfoutput=1

\documentclass[11pt]{article}

\usepackage[preprint]{acl}

\usepackage{times}
\usepackage{latexsym}

\usepackage[T1]{fontenc}

\usepackage[utf8]{inputenc}

\usepackage{microtype}

\usepackage{inconsolata}

\usepackage{graphicx}

\title{Luna: An Evaluation Foundation Model to Catch Language Model Hallucinations with High Accuracy and Low Cost}

\author{
    Masha Belyi$^*$ \quad \quad Robert Friel$^*$ \quad \quad Shuai Shao \quad \quad Atindriyo Sanyal \\ \\
    Galileo Technologies Inc. 
    \\ \texttt{\{masha,rob,ss,atin\}@rungalileo.io}
}

\usepackage{tabularx}
\usepackage{booktabs}
\usepackage{anyfontsize}
\usepackage{adjustbox}

\begin{document}
\maketitle
\def\thefootnote{*}\footnotetext{These authors contributed equally to this work}\def\thefootnote{\arabic{footnote}}

\begin{abstract}
Retriever-Augmented Generation (RAG) systems have become pivotal in enhancing the capabilities of language models by incorporating external knowledge retrieval mechanisms.
However, a significant challenge in deploying these systems in industry applications is the detection and mitigation of hallucinations—instances where the model generates information that is not grounded in the retrieved context.
Addressing this issue is crucial for ensuring the reliability and accuracy of responses generated by large language models (LLMs) in diverse industry settings.
Current hallucination detection techniques fail to deliver accuracy, low latency, and low cost simultaneously.
We introduce Luna: a DeBERTA-large (440M) encoder, fine-tuned for hallucination detection in RAG settings. We demonstrate that Luna outperforms GPT-3.5 and commercial evaluation frameworks on the hallucination detection task, with 97\% and 91\% reduction in cost and latency, respectively. Luna is lightweight and generalizes across multiple industry verticals and out-of-domain data, making it an ideal candidate for industry LLM applications.
\end{abstract}

\section{Introduction}

Large Language Models (LLMs) are broadly used in industry dialogue applications due to their impressive ability to hold a natural conversation and succeed on a variety of reasoning tasks \citep{zhao2023survey}. A key challenge in deploying customer-facing LLMs is their propensity for hallucinations, where the model presents cohesive, but factually incorrect information in conversation with a user \citep{roller-etal-2021-recipes, lin-etal-2022-truthfulqa}. Retrieval-augmented generation (RAG), a technique for incorporating knowledge relevant to each user query in the LLM prompt, effectively reduces LLM hallucinations in production systems \citep{lewis2020rag}. Yet, LLMs still often respond with nonfactual information that contradicts the knowledge supplied by RAG \cite{shuster-etal-2021-retrieval-augmentation, magesh2024hallucinationfree}.

\begin{figure}[t]
  \includegraphics[width=\columnwidth]{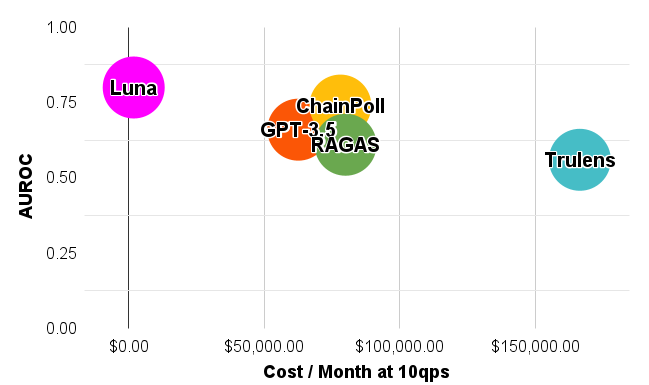}
  \caption{Luna is a lightweight DeBERTA-large encoder, fine-tuned for hallucination detection in RAG settings. Luna outperforms zero-shot hallucination detection models (GPT-3.5, ChainPoll GPT-3.5 ensemble) and RAG evaluation frameworks (RAGAS, Trulens) at a fraction of the cost and millisecond inference speed.}
  \label{fig:accuracy-vs-cost}
\end{figure}

Causes of hallucinations have been extensively studied across different LLM tasks \citep{zheng2024why, cao-etal-2022-hallucinated, das-etal-2022-diving}. Key contributing factors include knowledge cutoff \citep{vu2023freshllms}, randomness \citep{lee-2022-factuality-enhanced-llm}, faulty training data \citep{dziri-etal-2022-origin, lin-etal-2022-truthfulqa, mckenna2023sources}, and finetuning with large amounts of new knowledge \citep{gekhman2024does}. Apart from RAG, proposed mitigation solutions explore prompt engineering with chain of thought \citep{wei-2022-chain-of-thought}, finetuning \citep{zhang2024rtuning}, reinforcement learning with human feedback \citep{ouyang2022training-llms-to-follow-instructions-with-hf}, and specialized hallucination detection models \citep{wu2023ragtruth, lin-etal-2022-truthfulqa}.
For RAG specifically, evaluation frameworks like RAGAS \citep{es-etal-2024-ragas}, Trulens\footnote{https://www.trulens.org/}, and ARES \citep{saadfalcon2024ares} have emerged to offer automated hallucination detection at scale. 
However, these approaches rely on static prompts (RAGAS, Trulens) or finetuning on in-domain data (ARES), which limit their capacity to generalize to a breadth of industry applications.
\citet{gao-etal-2023-rarr} and \citet{wu2023ragtruth} take it a step further to successfully suppress hallucinations in LLM responses with a detect-and-replace technique. Though, due to prohibitively slow latency of their LLM evaluation models, real-time hallucination prevention in production systems still remains a challenge.

Customer-facing dialogue applications necessitate a hallucination detection system with high-accuracy, low cost, and low latency, such that hallucinations are caught and resolved before reaching the user. Few/zero-shot LLM approaches fail to meet the strict latency requirement due to model size. Moreover, though commericial LLMs like OpenAI's GPT models \citep{openai2023} achieve strong performance, querying customer data through 3rd party APIs is both costly and undesirable for privacy and security reasons. Finetuned BERT-size models can achieve competitive performance to LLM judges \citep{bohnet2023attributed, saadfalcon2024ares, gao-etal-2023-rarr, li2024attributionbench, yue-etal-2023-automatic}, offering lower latency and local execution. However, these models require annotated data for finetuning and have not been evaluated for large-scale, cross-domain applications.

In this paper, we introduce Luna - a lightweight RAG hallucination detection model that generalizes across multiple industry-specific domains and scales well for real-time deployment. Luna is a 440M parameter DeBERTa-large encoder that is finetuned on carefully curated real-world RAG data. From analysis of RAG in production settings, we identify long-context RAG evaluation as a previously unaddressed challenge and propose a novel solution that facilitates high precision long-context RAG hallucination detection. Through extensive benchmarking, we demonstrate that Luna outperforms zero-shot prompting and RAG evaluation frameworks on the hallucination detection task.

Our approach is closest to the concurrently proposed ARES automated RAG evaluation framework \citep{saadfalcon2024ares}, with a few key differences: (1) ARES requires a validation set of in-domain annotated data to finetune a custom evaluation model, while Luna is pre-trained on a cross-domain corpus for built-in generalization; (2) Luna accurately detects hallucinations on long RAG contexts; and (3) Luna is optimized to process up to 16k tokens in milliseconds on deployment hardware.


\section{Related Work}

\paragraph{Hallucination detection} Prior work on hallucination detection in natural language generation (NLG) is vast \citep{ji2023survey-hallucination}. SelfCheckGPT \citep{manakul-etal-2023-selfcheckgpt} and \citet{agrawal-etal-2024-language} are examples of heuristic consistency-based methods that detect unreliable LLM outputs by comparing multiple sampled responses from the same LLM. Others look to the internal state of the LLM, such as hidden layer activations \citep{azaria-mitchell-2023-internal} and token-level uncertainty \citep{varshney2023stitch} as a proxy signal for hallucinations. \citet{kadavath2022language} prompt the generating LLM to introspect and evaluate it's own responses. More generally, zero-shot \citep{es-etal-2024-ragas} and finetuned \citep{wu2023ragtruth, yue-etal-2023-automatic, muller-etal-2023-evaluating-attribution} LLM judges leverage LLM's inherent reasoning abilities to evaluate other LLM generations. Similarly, general purpose finetuned LLM evaluators \citep{kim2024prometheus} that have been shown to correlate with human judgements can also be applied to hallucination detection.

Our approach to finetune a small LM evaluator like in \citep{gao-etal-2023-rarr, saadfalcon2024ares} is the first to evaluate and optimize such a model for industry applications under strict performance, cost, and latency constraints.

\paragraph{NLI for closed-domain Hallucination Detection}
Existing research draws parallels between the hallucination detection task and the concept of entailment in Natural Language Inference (NLI). The goal of NLI is to determine the relationship between a premise and hypothesis, which can be one of: \textit{entailment}, \textit{contradiction}, or \textit{neutral}. In the past, NLI models have been used to evaluate factual consistency on closed-domain NLG tasks \citep{honovich-etal-2022-true-evaluating, dziri-etal-2022-evaluating}. The Attributable to Identified Sources (AIS) framework, introduced by \citet{rashkin2023measuringAttributionInNLG}, formally unifies the notions of factuality, attribution, hallucination, faithfulness, and groundedness - all terms used to measure the extent to which an LLM response is attributable to some source of ground truth. In followup work, NLI entailment has been shown to correlate with AIS scores \citep{gao-etal-2023-rarr, bohnet2023attributed, li2024attributionbench} and has become a standard baseline for AIS and hallucination detection models.

In this work, we use pre-trained NLI model weights as the starting point for Luna finetuning.




\section{Luna Model}

We fine-tune a DeBERTa-v3-Large \citep{he2023debertav} NLI checkpoint\footnote{https://huggingface.co/MoritzLaurer/DeBERTa-v3-large-mnli-fever-anli-ling-wanli} from \citet{laurer2022debertaNLI} with a shallow hallucination classifier on each response token. We train on the task of identifying \textit{supported} tokens in the response, given a query and retrieved context. Framing the problem in this way makes our work comparable to recent automated RAG evaluation efforts. Our definition of \textit{support} is synonymous with the \textit{answer faithfulness} metric explored in RAGAS \citep{es-etal-2024-ragas} and ARES \citep{saadfalcon2024ares}, Truelens \textit{groundedness}, and \textit{attribution} \citep{li2024attributionbench}. At inference, we treat spans with low support probabilities as hallucinated spans.  


Similar to \citet{gao-etal-2023-rarr} and \citet{wu2023ragtruth}, we aim to identify hallucinated spans in the response, rather than the less granular example-level hallucination boolean. While predicting spans is a more challenging task, it yields a more informative prediction to the end-user. Further, this approach sets us up for long-context prediction, which we discuss in detail next.


\begin{figure}[t]
  \includegraphics[width=\columnwidth]{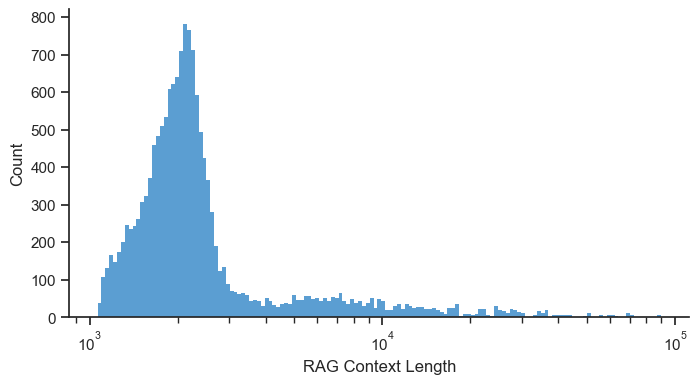}
  \caption{Distribution of RAG context token lengths in our QA RAG training split.}
  \label{fig:rag-context-length}
\end{figure}

\begin{figure*}[t]
  \includegraphics[width=\linewidth]{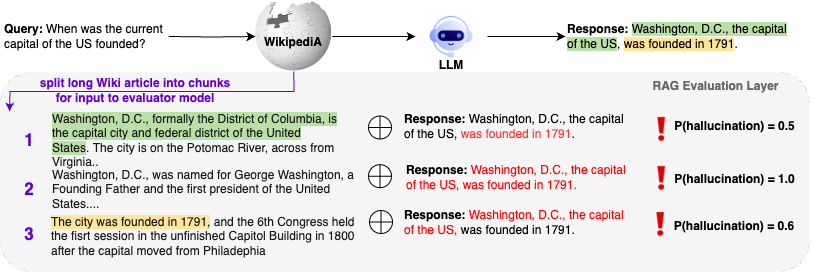}
  \caption{Long RAG context with naive chunking example. Naive context chunking leads to hallucination false positives when supporting information is scattered throughout the context. Without insight into which specific spans were suporrted/not supported by the context, it is impossible to arrive at the correct conclusion that the response in this example does NOT contain hallucinations.}
  \label{fig:long-context-challenge}
  
\end{figure*}
\begin{figure}[t]
  \includegraphics[width=\columnwidth]{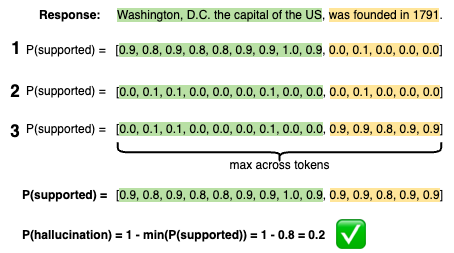}
  \caption{Illustration of Luna's token-level predictions for the example in Figure 3. Luna's token-level predictions are aggregated over context windows into a high-precision hallucination probability score.}
  \label{fig:long-context-solution}
\end{figure}

\subsection{Long Context RAG} In practice, we find that context length limitations are a significant pain point in industry applications. Custom RAG setups may retrieve a large number of context documents from various sources, or choose not to chunk the documents before passing them into the retriever. This results in long inputs to the RAG generator and evaluation models, sometimes even exceeding the token limit of select commercial LLMs. In Figure \ref{fig:rag-context-length} we visualize the context length distribution of our curated RAG dataset (detailed in Section \ref{section:training data}). While our base DeBERTa model can technically handle sequences of up to 24k \citep{he2021deberta}, computational complexity of transformer attention layers scale quadratically with input length. Moreover, though long-context LLMs like Claude-3 are becoming competitive on LLM leaderboards\footnote{https://huggingface.co/spaces/lmsys/chatbot-arena-leaderboard}, research shows that these models suffer from information loss \citep{liu2023lost} and may not be suitable for long-context RAG evaluation.

A naive solution is to chunk long-context RAG inputs into short segments and process them through the evaluator model in batches. Model predictions can then be aggregated over batch rows to predict example-level hallucination probabilities. Figure \ref{fig:long-context-challenge} illustrates how such chunking may result in false positives in cases where supporting information is scattered throughout the long context document(s). Instead, we leverage span-level predictions for a high-precision classifier over long sequence inputs.

\subsection{Long Context Chunking}
\label{section:long-context-chunking}
Consider a single input into the RAG evaluation model that consists of \textbf{C} context tokens $[c_1...c_C]$, \textbf{Q} question tokens $[q_1...q_Q]$, and \textbf{R} response tokens $[r_1...r_R]$. Assume we are working with an evaluator model that accepts maximum sequence length \textbf{L}, and  that \textbf{Q}+\textbf{R}<\textbf{L}, but \textbf{C} is much larger\footnote{the same approach easily extends to cases where R>L}. To fit the example into the model we break it up into windows of length \textbf{L}, such that each window contains the question, response, and a subset of the context tokens:
\begin{equation}
    w_i = [c_{i_1}...c_{i_l}] \oplus [q_1...q_Q] \oplus [r_1...r_R]
\end{equation}
where $l = L-Q-R$, and there are $\frac{N}{l}$ windows per example. In Figure \ref{fig:long-context-challenge} there are three such windows. Our model outputs support probabilities $p^i$ for each of the R response tokens in $w_i$ as:
\begin{equation}
    P_S(w_i) = [p_1^i...p_R^i]
\end{equation}

We train with a cross-entropy loss on each token output. During training, we leverage granular token-level support labels (Section \ref{section:data-labeling}) to adjust the training labels in each batch based on which context tokens are present in the window. For example, in Figure \ref{fig:long-context-challenge}, \texttt{"Washington, D.C., the capital of the US"} is supported in window 1, nothing is supported in window 2, and \texttt{"was founded in 1791"} is supported in window 3.

At inference, we aggregate example-level support probabilities by taking the token-level maximum over windows. Refer to Figure \ref{fig:long-context-solution} for an visual illustration of the steps described by equations 3-5 below. The example-level support probability for token j is defined as:
\begin{equation}
    p_j = \max_{1 \leq i \leq |w|}(p_j^i)
\end{equation}
where $|w| = \frac{N}{l}$ is the total number of windows we created in (1). To produce an example-level label, we take the minimum over R tokens:
\begin{equation}
    P_S = min(p_1...p_R)
\end{equation}
so that the overall support probability is no greater than the support probability of the least supported token in the response. Finally, we derive example hallucination probability $P\textsubscript{H}$ as 
\begin{equation}
    P_H = 1 - P_S
\end{equation}

\subsection{Training}
To leverage the full pre-trained NLI model, we initialize the hallucination prediction head with weights from the NLI classification head. The original NLI head is a 3-class single-layer perceptron with a neuron for each NLI class (entailment, contradiction, and neutral). During training, we optimize for low entailment probability and high contradiction probability for hallucinated tokens (and the opposite for supported tokens). At inference, we output the probability of entailment for each token.

We apply data transformation techniques to introduce additional variability for better generalization during training. Transformations include dropping and inserting context documents, and shuffling questions and responses between examples in batch. Training labels are adjusted accordingly with each transformation.

The model trains for 3 epochs with cross-entropy loss on the output of each response token. We initialize the learning rate to $5^{-6}$ for the base model layers and $2^{-5}$ for the classification head, and train with warmup and a linear decay rate. 

\begin{table}
    \begin{tabularx}{\columnwidth}{l|c|c|c|c}
        \toprule
        \textbf{Domain} & train & val  & test & \%H \\
        \hline
        customer support        & 4k    & 600   & 600   &  22\% \\
        finance                 & 38k   & 5k    & 5k    &  5\% \\
        biomedical research     & 22k   & 3k    & 3k    &  20\% \\
        legal                   & 1.5k  & 500   & 500   &  6\% \\
        general knowledge       & 9.5k  & 2k    & 2k    &  18\%  \\
        \bottomrule
    \end{tabularx}
    \caption{RAG QA data statistics. RAG context and questions are sourced from open-book QA datasets that cover five industry-specific domains. RAG responses are generated with GPT-3.5 and Claude-3-Haiku, and annotated with GPT-4-turbo. \textbf{\%H} indicates the fraction of hallucinated responses in each domain. }
    \label{table: training data stats}
\end{table}

\section{Data}
\subsection{RAG QA dataset}
\label{section:training data}
We recycle open-book QA datasets to construct a \textbf{RAG QA dataset}. Our goal is to simulate natural RAG examples that may occurr in production settings. We sample data from five industry verticals: customer support (DelucionQA \citep{sadat2024delucionqa}, EManual \citep{nandy-etal-2021-emanual}, TechQA \citep{castelli-etal-2020-techqa}), finance and numerical reasoning (FinQA \citep{chen-etal-2021-finqa}, TAT-QA \citep{zhu-etal-2021-tatqa}), biomedical research (PubmedQA \citep{jin-etal-2019-pubmedqa}, CovidQA \citep{moller-etal-2020-covidqa}), legal (Cuad \citep{hendrycks2021cuad}) and general knowledge (HotpotQA \citep{yang2018hotpotqa}, MS Marco \citep{nguyen2016msmarco}, HAGRID \citep{hagrid}, ExpertQA \citep{malaviya2024expertqa}). The combined dataset contains examples from a variety of difficult RAG task types, including numerical reasoning over tables, inference over multiple context documents, and retrieval from long contexts.  We reserve ~20\% of the dataset for validation and testing. Table \ref{table: training data stats} reports statistics of the data splits.


For each component dataset, we ignore the ground truth responses and generate two new responses per input with GPT-3.5 and Claude-3-Haiku. These models exhibit strong reasoning and conversational abilities \citep{chiang2024chatbot} at a low price point, which makes them realistic candidates for production RAG systems. We set temperature to 1 for generation to encourage diversity and potential hallucinations in the responses. Next, we describe how we annotate the data for training.


\subsection{Labeling}
\label{section:data-labeling}
We leverage GPT-4-turbo to annotate the RAG QA dataset. Refer to Section \ref{section:limitations} for a discussion on the limitations of this approach.

Before annotation, we split the context and response into sentences using nltk \citep{bird-loper-2004-nltk}. We pass the question along with the tokenized context and response sentences to GPT-4-turbo for annotation. For each sentence in the response, we instruct the LLM to identify which context sentences, if any, support the claim in the response. Tokens in sentences without any support are treated as hallucinations. We find that LLM responses often contain transition sentences and general statements that, while not supported by any specific context span, are generally grounded in the question and provided context. We instruct the annotator to label these as "generally supported", which we post-process to indicate support in every context window during training. Statements highlighting lack of sufficient information to answer the question also fall into this category.

We take measures to ensure high quality labels from our LLM annotator. First, we use chain-of-thought \citep{wei-2022-chain-of-thought}, which has been shown to increase agreement between LLM and human judgements \citep{he2024annollm}. Next, we request both response-level and sentence-level annotations that we compare to identify potentially noisy labels. For example, if GPT-4 claims a response as supported by the context as a whole, but identifies no supporting information for one or more claims in the response, we send the example for re-annotation. We re-annotate examples up to 3 times, after which <2\% of the data are still conflicting. After manual inspection, we find that the majority of the conflicts arise from partially supported sentences. Since our annotation scheme is binary on the sentence level (the full sentence is either supported or not), we resolve all tokens in partially supported sentences to "not supported" on both the sentence and example level. 





\begingroup
\renewcommand{\arraystretch}{1.1} 
\begin{table*}
  \centering
  \fontsize{7}{7}\selectfont
  \begin{tabularx}{\textwidth}{l|ccccccccc|ccc}
    \toprule
     Method & \multicolumn{3}{c}{\textsc{Question Answering}} & \multicolumn{3}{c}{\textsc{Data-to-Text Writing}} & \multicolumn{3}{c}{\textsc{Summarization}} & \multicolumn{3}{c}{\textsc{Overall}} \\
    \cmidrule(r){2-4}   \cmidrule(lr){5-7} \cmidrule(lr){8-10} \cmidrule(l){11-13}
     & Precision & Recall & F1 & Precision & Recall & F1 & Precision & Recall & F1 & Precision & Recall & F1 \\
    \midrule
    Prompt\textsubscript{gpt-3.5-turbo}$^\dagger$ & 18.8 & 84.4 & 30.8 & 65.1 & 95.5 & 77.4 & 23.4 & 89.2 & 37.1 & 37.1 & 92.3 & 52.9 \\
    Prompt\textsubscript{gpt-4-turbo}$^\dagger$ & 33.2 & 90.6 & 45.6 & 64.3 & 100.0 & 78.3 & 31.5 & 97.6 & 47.6 & 46.9 & 97.9 & 63.4 \\
    SelCheckGPT\textsubscript{gpt-3.5-turbo}$^\dagger$ & 35.0 & 58.0 & 43.7 & 68.2 & 82.8 & 74.8 & 31.1 & 56.5 & 40.1 & 49.7 & 71.9 & 58.8 \\
    LMvLM\textsubscript{gpt-4-turbo}$^\dagger$ & 18.7 & 76.9 & 30.1 & 68.0 & 76.7 & 72.1 & 23.2 & 81.9 & 36.2 & 36.2 & 77.8 & 49.4 \\
    Finetuned Llama-2-13B$^\dagger$ & 61.6 & 76.3 & \textbf{68.2} & 85.4 & 91.0 & \textbf{88.1} & 64.0 & 54.9 & \textbf{59.1} & 76.9 & 80.7 & \textbf{78.7} \\
    \midrule
    ChainPoll\textsubscript{gpt-3.5-turbo} & 33.5 & 51.3 & 40.5 & 84.6 & 35.1 & 49.6 & 45.8 & 48.0 & 46.9 & 54.8 & 40.6 & 46.7 \\
    RAGAS Faithfulness & 31.2 & 41.9 & 35.7 & 79.2 & 50.8 & 61.9 & 64.2 & 29.9 & 40.8 & 62.0 & 44.8 & 52.0 \\
    Trulens Groundedness & 22.8 & 92.5 & 36.6 & 66.9 & 96.5 & \underline{79.0} & 40.2 & 50.0 & 44.5 & 46.5 & 85.8 & 60.4 \\
    \midrule
    Luna & 37.8 & 80.0 & \underline{51.3} & 64.9 & 91.2 & 75.9 & 40.0 & 76.5 & \underline{52.5} & 52.7 & 86.1 & \underline{65.4} \\
    \bottomrule
  \end{tabularx}
  \caption{Response-level results on RAGTruth hallucination prediction task. Luna is compared against RAGTruth baselines reported in \citet{wu2023ragtruth} (rows marked with $^\dagger$), as well as our own baselines. RAGAS and Trulens are evaluation framewords that query GPT-3.5-turbo for hallucination detection. ChainPoll is our gpt-3.5-turbo ensemble prompt baseline. ChainPoll, RAGAS, Trulens, and Luna probability thresholds were tuned for best Overall F1. The top and second-best F1 scores are \textbf{bolded} and \underline{underlined}. Luna outperforms all prompt-based approaches and narrows the gap between other baselines and the 13B fine-tuned Llama, at a fraction of the cost.}
  \label{table:results-ragtruth-response-level}
\end{table*}
\endgroup

\begingroup
\renewcommand{\arraystretch}{1.1} 
\begin{table*}
  \centering
  \fontsize{7.5}{7.5}\selectfont
  \begin{tabularx}{\textwidth}{l|ccccc|c}
    \toprule
     Method & \textsc{Customer Support} & \textsc{Financial Reasoning} & \textsc{General Knowledge} & \textsc{Legal} & \textsc{Biomed} & \textsc{Overall} \\
    \midrule
    GPT-4-turbo annotator & 1.0 & 1.0 & 1.0 & 1.0 & 1.0 & 1.0 \\
    \midrule
    Prompt\textsubscript{gpt-3.5-turbo} & 0.68 & 0.67 & 0.67  & 0.63 & 0.64 & 0.66\\
    ChainPoll\textsubscript{gpt-3.5-turbo} & \textbf{0.76} & 0.74 & 0.75 & 0.71 & 0.71 & 0.74 \\
    \midrule
    RAGAS Faithfulness & 0.62 & 0.60 & 0.60 & 0.58 & 0.54 & 0.61\\
    Trulens Groundedness & 0.56 & 0.56 & 0.65 & 0.34 & 0.68 & 0.56\\
    \midrule
    Luna\textsubscript{in-domain} & \textbf{0.76} & \textbf{0.82} & \textbf{0.81} & \textbf{0.78} & \textbf{0.83} & \textbf{0.80}\\
    Luna\textsubscript{OOD} & 0.74 & 0.64 & - & 0.79 & - & - \\
    \bottomrule
  \end{tabularx}
  \caption{AUROC on the hallucination detection task on the RAG QA test set. Best score in each domain is \textbf{bolded}. Luna\textsubscript{in-domain} is our model trained on combined train splits from each domain.
  Luna\textsubscript{OOD} is the same model trained on a subset of General Knowledge and Biomed domains.
  }
  \label{table:results-ragbench-domains}
\end{table*}
\endgroup

\section{Evaluation}
\subsection{Datasets}
We evaluate Luna on a combination of existing academic benchmarks (RAGTruth) and real-world RAG data.

\paragraph{RAGTruth} RAGTruth is an expert-annotated corpus of 18k RAG examples with LLM-generated responses. The data are split into three RAG task types: Question Answering (QA), Data-to-text Writing, and News Summarization. Since Luna is only trained on QA RAG examples, we use this benchmark to evaluate our model's generalization to other RAG task types.

\paragraph{RAG QA Test Set} We also evaluate Luna on a held-out split of our RAG QA dataset (Section \ref{section:training data}). This serves as an in-domain test set for evaluating Luna performance across industry verticals.


\subsection{Baselines}
\paragraph{Zero-shot prompting} We evaluate GPT-3.5-turbo and GPT-4-turbo models from OpenAI as baselines. We prompt the LLMs to return an example-level boolean indicating whether or not a RAG response is supported by the associated RAG context. For RAGTruth we also include all baselines reported in the original paper.

\paragraph{Ensemble prompting} LLM ensembles have been shown to outperform single model judges by eliminating bias \citep{friel2023chainpoll, verga2024llm-juries}. We leverage ChainPoll \citep{friel2023chainpoll} with a chain-of-thought prompt for a stronger GPT-3.5-turbo baseline.

\paragraph{RAG Evaluation Frameworks} We evaluate two commercial RAG evaluation frmeworks: RAGAS (v0.1.7) \citep{es-etal-2024-ragas} and Trulens (v0.13.4). We report RAGAS Faithfulness and Trulens Groundedness metrics, which are designed for hallucination detection.


\subsection{Metrics}
For comparison with RAGTruth baselinse, we report best Precision, Recall, and F1 scores on RAGTruth. We tune model output probability thresholds for the best overall F1 and report all metrics at this optimal threshold. For other benchmarks, we report the area under the ROC curve (AUROC), which we consider a more informative metric that circumvents the need for threshold tuning.

\section{Results}
\label{section:results}
On the RAGTruth dataset, Luna outperforms all prompt-based approaches on the QA and Summarization tasks, and is competitive with GPT-3.5 evaluators on the Data-to-Text Writing task (Table \ref{table:results-ragtruth-response-level}). Overall, Luna is second only to the finetuned Llama-2-13B, which is expected given the significant difference in size between the two models (440M vs 13B). It's important to note that the Llama-2-13B baseline was trained on a subset of RAGTruth, as compared to Luna, which was trained on a QA-only dataset with a different data distribution. Nevertheless, we find that Luna generalizes well to the out-of-domain task types. Additionally, the gains in cost and inference speed we achieve with the lightweight Luna model (Sections \ref{section:cost-discussion},  \ref{section:latency-discussion}) offset the performance gap.

Results on the RAG QA test set are reported in Table \ref{table:results-ragbench-domains} and follow a similar pattern. Luna outperforms the baselines across all verticals.

We also evaluate the model's cross-domain generalization by training on a subset of General Knowledge and Biomedical Domains, and evaluating on the others. We refer to this model as Luna\textsubscript{OOD}. We find that Luna\textsubscript{OOD} still outperforms most baselines on the out-of-domain subsets. However, generalization to the Financial Reasoning domain is weak. Examples in this domain require reasoning over tabular data, which Luna\textsubscript{OOD} never observes in training. Fine-tuning on the Financial Reasoning domain greatly boosts performance, increasing AUROC from 0.64 to 0.82. 


\begin{table}
    \fontsize{10}{10}\selectfont
    \begin{tabularx}{\columnwidth}{l|ccc}
        \toprule
            & 0-5k & 5k-16k & 16k+ \\
         (count in test) & (223) & (209) & (78)\\
        \midrule
        Prompt\textsubscript{gpt-3.5-turbo}     & 0 & -12.11\% & -100\%\\
        ChainPoll\textsubscript{gpt-3.5-turbo}  & 0 & -8.97\% & -100\%\\
        \midrule
        RAGAS Faithfulness                      & 0 & -4.36\% & -100\%\\
        Trulens Groundedness                    & 0 & -6.38\% & -100\%\\
        \midrule
        Luna                                   & 0 & -12.55\% & -31.98\% \\
        Luna\textsubscript{example}             & 0 & -21.44\% & -43.75\% \\
        \bottomrule
    \end{tabularx}
    \caption{Relative hallucination detection performance of various models on shor(0-5k), medium(5k-16k), and long(16k+) context lengths. \textbf{Luna} is our best fine-tuned DeBERTA-large model, and \textbf{Luna\textsubscript{example}} is a version of Luna that makes hallucination predictions at example level. All GPT-3.5-based baselines (including RAGAS, Trulens) fail on input lengths >16k, while Luna maintains 88\% and 68\% of its's performance on medium (5k-16k) and long (16k+) context lengths, respectively. Luna\textsubscript{example} also struggles more with long context lengths that Luna.}
    \label{table:results-context-length}
\end{table}

\section{Discussion}

\subsection{Long Context Hallucination Detection}
In Table \ref{table:results-context-length} we report Luna's performance against baselines on a range of RAG context lengths. For this analysis we sample data from CUAD \citep{hendrycks2021cuad}, one of the RAG QA component datasets, which passes full-length legal contracts as context inputs into RAG. This dataset contains the largest range of context lengths in RAG QA.

We find that performance of all models inversely correlates with context length. However, while the GPT-3.5-powered baselines fail completely at  the GPT-3.5 context limit (16k tokens), Luna maintains 68\% of it's performance on that subset.



To validate the efficacy of our span-level prediction and long context chunking approach (Section \ref{section:long-context-chunking}), we do an ablation study where we compare our best model to a version of Luna that makes example level predictions, referred to as Luna\textsubscript{example} in Table \ref{table:results-context-length}. As shown in Figure \ref{fig:long-context-challenge}, we expect Luna\textsubscript{example} to perform worse on long contexts. Our findings confirm this hypothesis: although the hallucination detection performance of both Luna and Luna\textsubscript{example} degrades with increasing context lengths, Luna\textsubscript{example} exhibits a greater degradation than Luna.

\subsection{Cost vs Accuracy Trade-offs}
\label{section:cost-discussion}
API-based hallucination detection methods accrue substantial costs if used continuously in production settings. Luna outperforms GPT-3.5-based approaches while operating at a fraction of the cost. In Figure \ref{fig:accuracy-vs-cost} we illustrate the trade-off between monthly maintenance costs and accuracy for Luna versus our GPT-3.5-based baselines. Costs are estimated assuming average throughput of 10 queries per second, with average query length of 4000 tokens. We use OpenAI API\footnote{https://openai.com/api/pricing/} and AWS cloud\footnote{https://aws.amazon.com/ec2/pricing/on-demand/} pricing at the time of writing. Detailed cost calculations can be found in Appendix \ref{appendix:cost-calculations}. 

Although we do not explicitly compare pricing against larger fine-tuned models such as Llama-2-13B, we note that hosting a multi-billion parameter model demands substantially more compute resources than Luna, which would be reflected in the overall cost.

\subsection{Latency Optimizations}
\label{section:latency-discussion}
We optimize Luna and its deployment architecture to process up to 16k input tokens in under one second on NVIDIA L4 GPU. To achieve this, we deploy an ONNX-traced model on NVIDIA Triton server with TensorRT backend. We leverage Triton's Business Logic Scripting (BLS) to optimize the data flow and orchestration between GPU and CPU resources. BLS intelligently allocates resources based on the specific requirements of each inference request, ensuring that both GPU and CPU are utilized effectively and that neither resource becomes a bottleneck. We also tune our inference model maximum input length for optimal performance. While increasing the maximum sequence length would reduce the size and number of batches processed by the model (see Section \ref{section:long-context-chunking}), transformer layer computational complexity also scales quadratically with input length. We determine token length of 512 to be the most effective. Finally, we optimize pre-and post-processing python code for maximum efficiency. Table \ref{table:latency-optimization} in Appendix details the latency reductions achieved at each optimization step.



\section{Conclusion}
In this work we introduced Luna: a cost-effective hallucination detection model with millisecond inference speed. Luna eliminates dependency on slow and expensive 3rd party API calls, and enables practitioners to effectively address hallucinations in production. The proposed model can be hosted on a local GPU, guaranteeing privacy that 3d-party API's cannot.

\subsection{Limitations}
\label{section:limitations}
\paragraph{Closed Domain Hallucinations} Luna's efficacy is limited to closed domain hallucination detection in RAG settings. Due to its size, Luna lacks the necessary world knowledge to detect open domain hallucinations. For open-domain applications, Luna relies on a high-quality RAG retriever to provide the necessary context knowledge for an input query.

\paragraph{LLM Annotations} 
LLM's remarkable zero-shot abilities have encouraged researchers to consider LLMs for annotation and synthetic data generation. Replacing human annotators with LLMs offerst substantial efficiency and cost savings \citep{wang-etal-2021-want-reduce}.
However, LLM performance on various annotation tasks is still controversial, with some studies reporting high correlations between LLM and human judgements \citep{chiang-lee-2023-large, he2024annollm, verga2024llm-juries}, while others advise caution \citep{li-etal-2023-synthetic, wang2024chatgpt}.

In this work, we recognize the potential noise and bias introduced in our training and evaluation data by automated GPT-4-turbo annotations. We hypothesize that our model derives greater advantages from training on a large-scale dataset, facilitated by low-cost LLM annotation, than it is hindered by potential noise within the data. After taking steps to ensure annotation quality (Section \ref{section:data-labeling}), we observe competitive performance on RAGTruth, a human-annotated benchmark in Section \ref{section:results}. This evaluation provides external validation for our model outputs, although we acknowledge that performance could potentially be enhanced with higher quality annotation sources.

\paragraph{Sentence-level annotations} Luna is trained on sentence-level annotations, i.e. there is an assumption that a sentence is either supported or not supported. This is most often the case, but future work can explore token-level labels for compound sentences with partially supported claims.


\subsection{Future Work}
Hallucinations in RAG output highlight weaknesses of the generator model. However, it is equally important to consider the quality of the retriever and its contribution to the overall performance of a RAG system. A sub-optimal retriever may supply irrelevant context to the generator, making it difficult for the generator to produce an accurate response. A comprehensive RAG evaluation model should therefore assess all dimensions of the RAG system. To this end, metrics like \textit{context relevance} have been explored to assess the quality of retrieved RAG contexts \citep{es-etal-2024-ragas, saadfalcon2024ares}.

In future work, we propose to leverage Luna for measuring a comprehensive suite of RAG metrics. One cost-effective approach could be to augment the current DeBERTA architecture with additional prediction heads that output multiple metrics in one forward pass. We hypothesize that the shared weights of the base encoder layers may enhance the performance of each head.

\bibliography{Luna}

\appendix

\section{Response Generation Prompt}
\label{appendix-response-generation-prompt}
We use the following prompt template to generate LLM responses for each sample in our QA RAG dataset. Context documents, separated by line breaks, along with the question are slotted in for each generation sample.
\begin{quote}
Use the following pieces of context to answer the question.

\{documents\}

Question: \{question\}
\end{quote}

\section{Cost Calculations}
\label{appendix:cost-calculations}
Costs are estimated assuming average throughput of 10 queries per second (qps), with average RAG query length of 4000 tokens, and NVIDIA L4 GPU deployment hardware. When estimating LLM cost for >1qps we assume concurrency is implemented to process multiple queries in parallel.

\paragraph{Luna Costs} Empirically, we find that each L4 can serve up to 4qps. At the time of writing, the monthly cost of running a g6.2xlarge GPU instance on AWS cloud is \$700\footnote{https://aws.amazon.com/ec2/pricing/on-demand/}. Thus, we estimate total monthly cost for 10qps throughput as
\begin{equation}
    \$700 * \frac{10}{4} = \$1750
\end{equation}

\paragraph{OpenAI Costs} At the time of writing, querying GPT-3.5-turbo through OpenAI API costs \$0.50 / 
1M input tokens and \$1.50 / 1M output tokens\footnote{https://openai.com/api/pricing/}. In our test set, we observe the average output token length from GPT-3.5 at 200 tokens. Using average input length of 4000 tokens, the cost of a single query is roughly
\begin{equation}
    (4k*\$0.5+ 200*\$1.5)/1M = \$0.0023
\end{equation}
Using 2,592,000 seconds/month, the monthly cost of serving 10qps with GPT-3.5 is:
\begin{equation}
    10qps * 2,592,000 * \$0.0023 = \$59,616
\end{equation}

With ChainPoll ensemble, we request 3 outputs per query, bringing the cost of a single query up to
\begin{equation}
    (4k*\$0.5+ 3*200*\$1.5)/1M = \$0.0029
\end{equation}
And the total monthly cost for 10qps to:
\begin{equation}
    10qps * 2,592,000 * \$0.0029 = \$75,168
\end{equation}

\paragraph{RAGAS Costs}
RAGAS makes 2 OpenAI API calls per an input RAG example. The first query extracts a list of claims from the response. The second requests the LLM to evaluate the faithfulness of each extracted claim to the RAG context. We estimate that the output length of the first query is roughly equal to the length of the RAG response; and the output length of the second query is roughly 3x the length of the response, since it includes the original claims followed by a faithfulness score and an explanation. Factoring in overhead token length of each prompt, we calculate the cost per query to be
\begin{equation}
    Query 1 = \$380 / 1M 
\end{equation}
\begin{equation}
    Query 2 = \$2730 / 1M
\end{equation}
Then, the monthly cost of serving 10qps is:
\begin{equation}
    10qps * 2,592,000 * (\$380 + \$2730)/1M = \$79,937
\end{equation}

\paragraph{Trulens Costs}
Trulens makes 1 OpenAI per each sentence in the response. For this calculation, we estimate 3 sentences per response, which aligns with our obesrvations on the QA RAG dataset. Each query returns original sentence, a groundedness score (1-10), and an explanation. Here we assume that the token length of the explanation is roughly equal to the token length of the input sentence. The cost of a single query is roughly
\begin{equation}
    (4k*\$0.5+ 2*75*\$1.5)/1M = \$0.0022
\end{equation}
Using 2,592,000 seconds/month, the monthly cost of serving 10qps with Trulens is:
\begin{equation}
    10qps * 2,592,000 * 3 * \$0.0022 = \$173,016
\end{equation}

\section{Latency Optimizations}
\label{appendix:latency-optimization}
We optimize Luna and its deployment architecture to process up to 16k input tokens in under one second on NVIDIA L4 GPU. Table \ref{table:latency-optimization} details the latency reductions and how they were achieved.

\begin{table}[t]
    \begin{tabularx}{\columnwidth}{Xc}
        \toprule
        Optimization & s/16k \\
        \midrule
        baseline                            & 3.27 \\
        TensorRT backend                    & 2.09 \\
        efficient pre- and post- processing code      & 1.79 \\
        512 max model length                & 0.98 \\
        BLS                                 & 0.92 \\
        \bottomrule
    \end{tabularx}
    \caption{Impact of latency optimizations on Luna inference speed. Reporting inference speed in seconds for processing 16k input tokens.}
    \label{table:latency-optimization}
\end{table}

\section{Latency Comparison}
We empirically estimate the latency of Luna and each baseline model. Luna latency is discussed in Appendix \ref{appendix:latency-optimization}. For LLm models that query OpenAI API, we calculate the average latency per query after querying the API multiple times with an input of 4000k tokens, split between 3800 tokens for the context, 25 tokens for the question, and 75 tokens for the response.

\begin{table}[hb]
    \begin{tabularx}{\columnwidth}{Xcc}
        \toprule
        Model & s/4k & \%change \\
        \midrule
        Luna               & 0.23 & -       \\
        GPT-3.5            & 2.5 & -91\%    \\
        ChainPoll n=3      & 3.0 & -93\%    \\
        Trulens            & 3.4 & -93\%    \\
        RAGAS              & 5.4 & -96\%      \\
        \bottomrule
    \end{tabularx}
    \caption{Model latency (in seconds), comparing Luna to LLM baselines. We also report the \% difference between Luna and LLM-based models.}
    \label{table:latency-comparison}
\end{table}


\end{document}